\documentclass[letterpaper]{article} 
\usepackage{aaai2027}  
\usepackage[hyphens]{url}  
\usepackage{graphicx} 
\usepackage{natbib}  
\usepackage{caption} 
\usepackage{algorithm}
\usepackage{algorithmic}

\usepackage{multirow}
\usepackage{pifont}
\usepackage{enumitem}
\usepackage{tcolorbox}
\usepackage{adjustbox}
\usepackage{subcaption}

\newcommand{\cmark}{\ding{51}} 

\usepackage{newfloat}
\usepackage{listings}
\DeclareCaptionStyle{ruled}{labelfont=normalfont,labelsep=colon,strut=off} 
\floatstyle{ruled}
\newfloat{listing}{tb}{lst}{}
\floatname{listing}{Listing}

\usepackage{booktabs}

\title{Net-Reason: Benchmarking Accuracy and Token Cost on Table Reasoning for Network-Operations Agents}

\author {
    Yannis Belkhiter\textsuperscript{\rm 1,\rm 2},
    Seshu Tirupathi\textsuperscript{\rm 1},
    Giulio Zizzo\textsuperscript{\rm 1},
    John D. Kelleher\textsuperscript{\rm 2},
    Merim Dzaferagic\textsuperscript{\rm 2}
}
\affiliations {
    \textsuperscript{\rm 1}IBM Research, \textsuperscript{\rm 2}Trinity College Dublin\\
    yannis.belkhiter@ibm.com, seshutir@ie.ibm.com, giulio.zizzo2@ibm.com, john.kelleher@tcd.ie, merim.dzaferagic@tcd.ie
}

\begin{document}

\maketitle

\begin{abstract}
Network-operations agents need to interpret technical evidence before recommending or executing changes. Reasoning over tables in standards, technical guides, and configuration references is one component of this capability, yet its accuracy and inference-time efficiency require targeted evaluation. We introduce Net-Reason, a table question-answering benchmark containing 503 human-verified questions grounded in 36 network technical documents of six source types. The benchmark covers four question categories: data extraction, aggregation, logic, and multi-row/multi-column reasoning. We evaluate five open-weight models (1.5-32B) under seven inference configurations (including base, instruct and reasoning). Reasoning variants improve mean G-Eval correctness scores over base counterparts by approximately 13–42 percentage points, while generating 2–15 times as many output tokens. Aggregation and multi-row/multi-column questions generally yield lower scores and require more tokens than data extraction and logic questions. 
Agentic pipelines based on ReAct and Plan-Execute-Verify exhibit similar trade-offs, although their effectiveness is model-dependent. At least one of the two pipelines improves mean correctness scores for four of the six evaluated models, with gains of up to nine percentage points, while generating 1.5--40 times as many output tokens.
We further evaluate existing early-stopping methods and find that their answer-quality and token-saving trade-offs vary by question category. 
Net-Reason provides a controlled testbed for evaluating the trade-offs between correctness and token efficiency across model-level and pipeline-level reasoning approaches relevant to network operations.
\end{abstract}

\begin{links}
    \link{Code}{https://anonymous.4open.science/r/Net-Reason-82C7/}
\end{links}

\section{Introduction}

Large Language Model (LLM)-based agents are increasingly being investigated for network management, configuration, and troubleshooting \cite{boateng2024surveylargelanguagemodels,miao2023empiricalstudynetopscapability}. In an agentic network-operations loop, an agent must interpret network observations, retrieve relevant technical evidence, reason about possible interventions, validate proposed actions before execution, and monitor their effects to inform subsequent decisions. The correctness of these interventions therefore depends not only on the agent's ability to interact with the network, but also on its ability to interpret the evidence that informs its decisions.


\begin{table*}[h]
\centering
\tiny
\caption{Overview of the literature. \emph{Type of reasoning} includes works that explicitly categorize the different types of reasoning involved in the questions (not only by difficulty, e.g. Easy/Medium/Hard). \emph{Multi doc. type} include benchmarks that contain different type of documents as part of the context of the questions. \emph{Eff. analysis} refers to the joint analysis of correctness and output-token use.} 
\label{tab:existing-benchmarks}
\begin{tabular}{ccccccccccc}
    \toprule
    \multirow{2}{*}{\textbf{Datasets}} & \multirow{2}{*}{\textbf{Ref.}} & \multirow{2}{*}{\textbf{Domain}} & \multirow{2}{*}{\textbf{Size}} & \multicolumn{2}{c}{\textbf{Question complexity}} & \multicolumn{2}{c}{\textbf{Domain-specific}} & \multirow{2}{*}{\textbf{Reasoning type}} & \multirow{2}{*}{\textbf{Multi doc. type}} & \multirow{2}{*}{\textbf{Eff. analysis}} \\
    
    \cmidrule(lr){5-6} \cmidrule(lr){7-8} 
    
     & & & & Retrieval & Reasoning & Tech. doc. & Knowl. & & \\
    \midrule
    WikiTableQuestion & \cite{pasupat2015compositionalsemanticparsingsemistructured} & General &  22,033 &  & \cmark \\
    \midrule
    WikiSQL & \cite{zhong2017seq2sqlgeneratingstructuredqueries} & General & 80,654 & \cmark & \cmark & & &  \\
    \midrule
    SQA & \cite{iyyer-etal-2017-search} & General & 17,553 & \cmark & & & &  \\
    \midrule
    Open-WikiTable & \cite{kweon2023openwikitabledatasetopendomain} & General & 67,023 & \cmark & \cmark & & & &  \\
    \midrule
    E2E-WTQ & \cite{pan2021cltrendtoendtransformerbasedcell} & General & 1,216 & \cmark & & & & & \cmark \\
    \midrule
    E2E-GNQ & \cite{pan2021cltrendtoendtransformerbasedcell} & General & 789 & \cmark & & & & & \cmark \\
    \midrule
    NQ-table & \cite{herzig2021opendomainquestionanswering} & General & 11,628 & \cmark & & & & \\
    \midrule
    MiMoTable & \cite{li2024mimotablemultiscalespreadsheetbenchmark} & General & 1,719 & \cmark & \cmark & & & \cmark & \\
    \midrule
    MMQA & \cite{talmor2021multimodalqacomplexquestionanswering} & General & 29,918 & & \cmark & & & \cmark & \cmark \\
    \midrule
    FREB-TQA & \cite{zhou2024frebtqafinegrainedrobustnessevaluation} & General & 8,590 & \cmark & \cmark \\
    \midrule
    \midrule
    TAT-QA & \cite{zhu2021tatqaquestionansweringbenchmark} & Finance & 16,552 & \cmark & \cmark & \cmark & \cmark & \cmark &  \\
    \midrule
    FinQA & \cite{chen2022finqadatasetnumericalreasoning} & Finance & 8,281 &  & \cmark & \cmark & \cmark &  &  \\
    \midrule
    AIT-QA & \cite{katsis2021aitqaquestionansweringdataset} & Airline & 515 &  & \cmark & \cmark & \cmark &  &  \\
    \midrule
    TeleQnA & \cite{maatouk2023teleqnabenchmarkdatasetassess} & Network & 10,000 & \cmark & & \cmark & \cmark \\
    \midrule
    TeleTables & \cite{ezzakri2025teletablesbenchmarklargelanguage} & Network & 500 & \cmark & \cmark & \cmark & \cmark &  &  \\
    \midrule
    TeleMath & \cite{colle2025telemathbenchmarklargelanguage} & Network & 500 & & \cmark &  & \cmark & \cmark  \\
    \midrule
    TeleLogs & \cite{sana2025reasoninglanguagemodelsroot} & Network & - & & \cmark &  & \cmark & &  \\
    \midrule
    3GPP TSG & \cite{nikbakht2024tspecllmopensourcedatasetllm} & Network & 100 & \cmark & \cmark & \cmark & \cmark & \\
    \midrule
    ORANBench & \cite{gajjar2024oranbench13kopensourcebenchmark} & Network & 13,952 & & & \cmark & \cmark \\
    \midrule
    \textbf{Net-Reason} (Ours) & - & Network & 503 & \cmark & \cmark & \cmark & \cmark & \cmark & \cmark & \cmark \\
    \bottomrule
\end{tabular}
\end{table*}

A significant part of this evidence is structured in tables within network standards, technical guides, and configuration references \cite{3gpp_ts23501,3gpp_ts22261,srsran_config}. Consider an agent tasked with identifying a suitable Quality of Service (QoS) configuration for a service with specific latency and reliability requirements. The agent must identify candidate 5G QoS Identifiers (5QIs), compare their packet-delay budgets and packet-error rates, and determine their associated resource types \cite{3gpp_ts23501}. While individual parameters can be retrieved through simple lookups, identifying suitable candidates requires combining information across multiple rows and columns. An incorrect interpretation may lead to an unsuitable configuration recommendation, even when the agent correctly retrieves the relevant evidence and invokes the appropriate network tools.

The reasoning capabilities required within such a loop vary with the operational task. Retrieving an individual configuration parameter may require only a single model invocation, whereas evaluating multiple configuration alternatives against several constraints may benefit from multi-step reasoning, external computation, and verification. However, additional reasoning steps and tool interactions increase inference overhead, which is particularly relevant when agents operate repeatedly or under operational time constraints. Determining when additional reasoning is beneficial is therefore important for the design of efficient network-operations agents.

Table Question Answering (TQA) provides an established setting for evaluating the ability of language models to interpret structured information \cite{zhou2025tablequestionansweringera}. Existing telecommunications benchmarks, such as TeleTables \cite{ezzakri2025teletablesbenchmarklargelanguage}, have extended this evaluation to network-specific technical standards. However, the trade-offs between answer correctness and inference overhead across model-level and pipeline-level reasoning approaches remain underexplored in the networking domain. Understanding how these trade-offs vary across models and table operations is essential for selecting appropriate reasoning strategies within network-operations agents.

In this paper, we introduce Net-Reason, a benchmark for evaluating the correctness and token efficiency of reasoning over network technical tables. It contains 503 question--answer pairs spanning four categories: data extraction, aggregation, logic, and multi-row/multi-column reasoning. Each question is associated with a specific table supplied as context, allowing us to evaluate evidence interpretation independently of document retrieval. A no-table ablation assesses the contribution of the supplied evidence to answer correctness. Net-Reason therefore evaluates a specific capability within an agentic network-operations loop rather than the complete process of observation, decision-making, and execution.

We use Net-Reason to investigate how answer quality and output-token use vary across model configurations, agentic pipelines, and question categories. Across the evaluated models, reasoning variants achieve higher correctness scores than their base counterparts, but generate substantially more tokens. Agentic pipelines further increase token consumption relative to direct inference, but their benefits are model-dependent. We further evaluate whether existing early-stopping methods can reduce generated tokens while preserving answer quality. The resulting trade-offs vary across question categories.
Our contributions are: 

\begin{itemize}[leftmargin=*, itemsep=0pt, topsep=0pt]

    \item \textbf{Net-Reason.} We introduce a telecommunications TQA benchmark 
    comprising 503 question--answer pairs across four reasoning categories and six technical source types. We further assess dataset quality and the reliability of the evaluation metric through independent human audits.

    \item \textbf{Model-level reasoning.} We evaluate five open-weight models under seven inference configurations, including base, instruct, and reasoning variants. Reasoning variants achieve the highest correctness scores across the five evaluated models, at 2--15 times the output-token cost, with gains and costs that depend on question category.
    
    \item \textbf{Pipeline-level reasoning.} We evaluate ReAct and Plan-Execute-Verify against direct inference across six LLMs. At least one of the two pipelines improves mean correctness scores for four of the six models, with gains of up to nine percentage points and 1.5--40 times as many output tokens.
    
    \item \textbf{Inference-time efficiency.} We evaluate three classes of early-stopping methods and show that their answer-quality and token-saving trade-offs depend on the question category.

\end{itemize}

\begin{figure*}[h]
    \centering
    \includegraphics[width=0.95\linewidth]{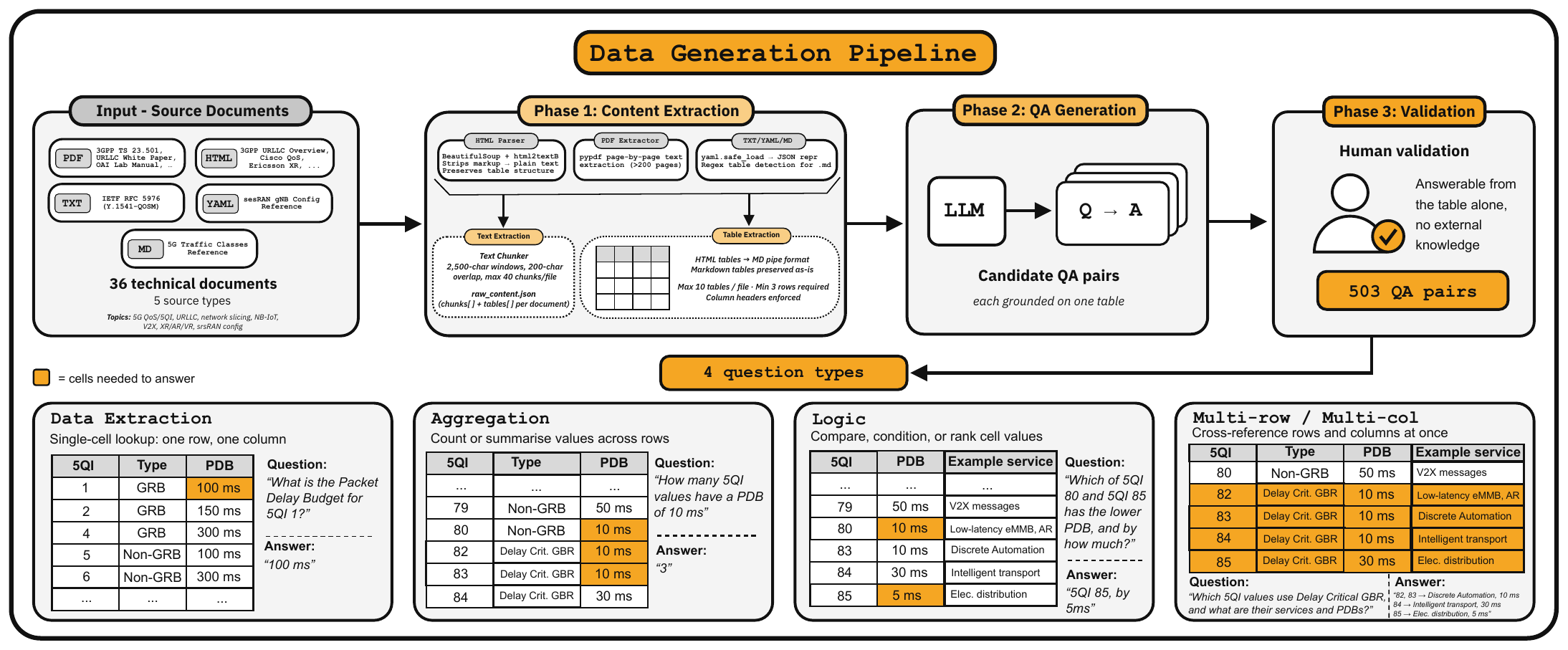}
    \caption{Data generation pipeline.}
    \label{fig:pipeline}
\end{figure*}

\section{Related work}

\textbf{General TQA.} The field of Table Question Answering is expanding quickly \cite{zhou2025tablequestionansweringera}. Table \ref{tab:existing-benchmarks} summarizes existing benchmarks. However, most existing benchmarks are domain-agnostic and do not evaluate models on real-world environments.

\noindent \textbf{TQA specific to Communication Networks.} Recently, a growing body of work suggested benchmarks to evaluate LLMs on the Network domain. TeleQnA \cite{maatouk2023teleqnabenchmarkdatasetassess} is the first QA dataset to evaluate a model's knowledge over 3GPP standards, but focuses on information retrieval rather than reasoning tasks. Other benchmarks, such as $\tau^2$-Bench telecom \cite{barres2025tau2benchevaluatingconversationalagents}, evaluates agents in a troubleshooting environment. TeleMath \cite{colle2025telemathbenchmarklargelanguage} evaluates mathematical reasoning but does not operate over technical documents. Closer to our work, TeleTables \cite{ezzakri2025teletablesbenchmarklargelanguage} targets table reasoning over telecom documents. However, it is restricted to a specific type of technical documents (3GPP specifications), and does not categorize questions by table operation or analyse inference costs.

\section{Net-Reason dataset} 
\label{sec:dataset_generation}

This section presents an overview of the Net-Reason dataset. We first detail the source and processes to generate and validate the data. We then describe the dataset obtained.

\noindent \textbf{Source documents.} Net-Reason is built on $36$ technical documents spanning a vertical slice of the 5G knowledge stack: from chip-level timing parameters (srsRAN) to network-architecture principles (3GPP TS 23.501) to end-to-end service requirements (AR/VR latency, V2X reliability). This corpus includes $6$ type of source documents, namely: 3GPP specifications, IETF and ITU-T standards, industry white papers and technology reviews, research papers, online reference portals, and configuration references. A full list of source documents is available in Appendix A.

This breadth requires models to reason across abstraction levels rather than exploiting vocabulary patterns from a single source. The coexistence of formal standards and practitioner guides also creates genuine terminological variation: the same concept can appear under different names and in different units across documents, which is a non-trivial disambiguation challenge for tabular tasks.

\noindent \textbf{Generation and validation pipeline.}
Questions are generated through an automated 3-phases pipeline (see Figure \ref{fig:pipeline}):

\begin{enumerate}[leftmargin=*, itemsep=0pt, topsep=0pt]

    \item \textbf{Content extraction.} We first parse the technical documents (in HTML, PDF, YAML, and Markdown original source) to produce semi-structured tables and text.

    \item \textbf{QA generation.} We then prompt an LLM to generate questions on each tables. Each question is grounded on the Markdown table.

    \item \textbf{Validation.} Finally, we validated each QA sample with human feedback. We ensure that each question-answer pair is answerable solely from the provided table, without relying on external knowledge.
    
\end{enumerate}
An initial set of 510 question--answer pairs underwent systematic quality filtering to assess answerability, technical relevance, question clarity, and consistency with the assigned question category. Of these, 403 samples were retained unchanged, 16 were revised, and 91 were discarded. To compensate for the discarded samples, we generated 85 additional questions and removed one near-duplicate, resulting in a final dataset of 503 question--answer pairs.

\noindent Our process resulted in $503$ tabular human-verified QA pairs, each traceable to a specific table from a source document. 
Figure \ref{fig:net-reason-analysis} in Appendix B shows statistics of Net-Reason.
We additionally audited a random subset with three independent expert annotators. 

\begin{figure*}[h]
    \centering
    \begin{subfigure}{0.22\linewidth}
        \includegraphics[width=1\linewidth]{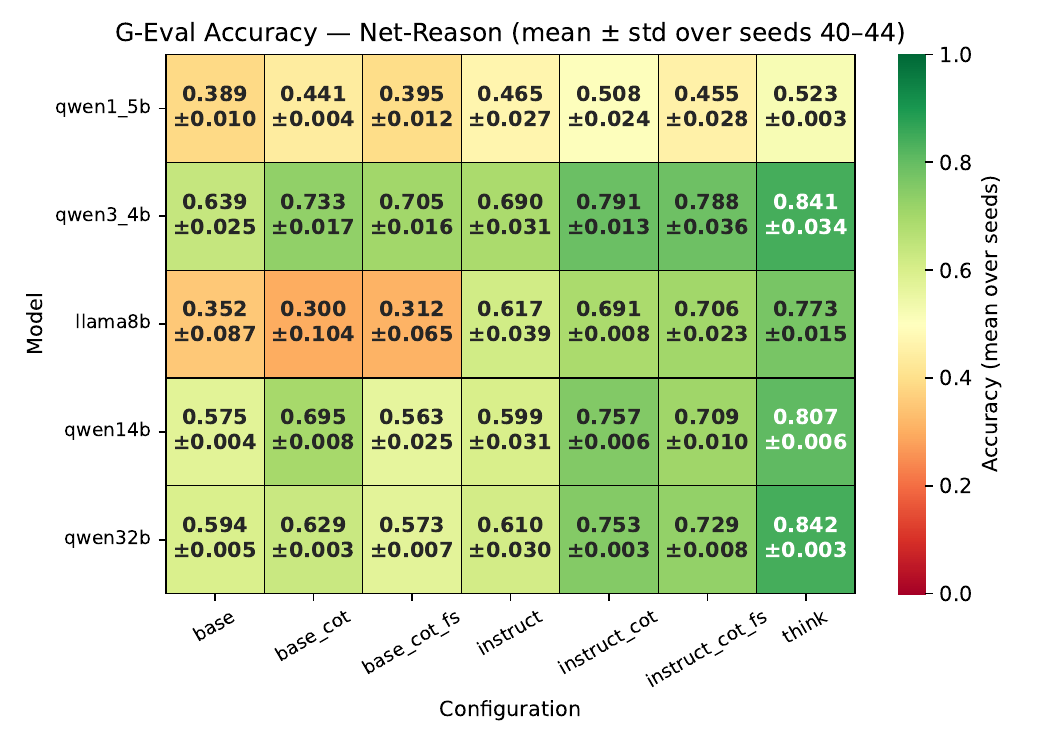}
        \caption{Overall Accuracy}
        \label{fig:heatmap-overall}
    \end{subfigure}
    \hfill
    \begin{subfigure}{0.22\linewidth}
        \includegraphics[width=1\linewidth]{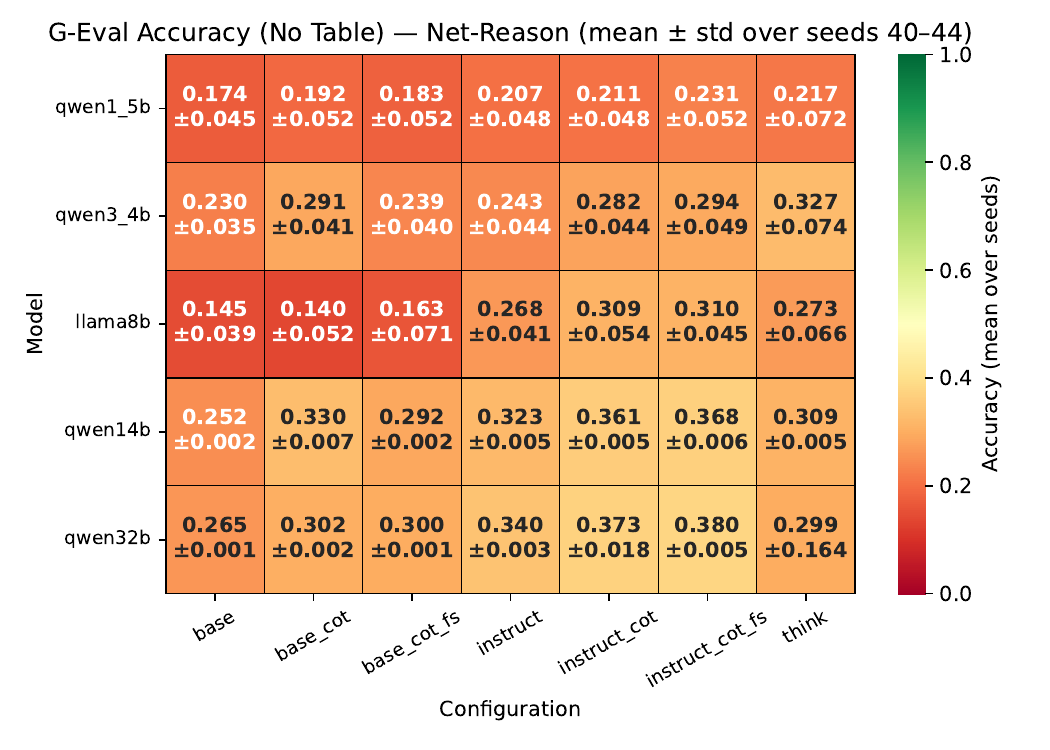}
        \caption{Removing Tables}
        \label{fig:heatmap-overall-no-table}
    \end{subfigure}
    \hfill
    \begin{subfigure}{0.28\linewidth}
        \includegraphics[width=1.0\linewidth]{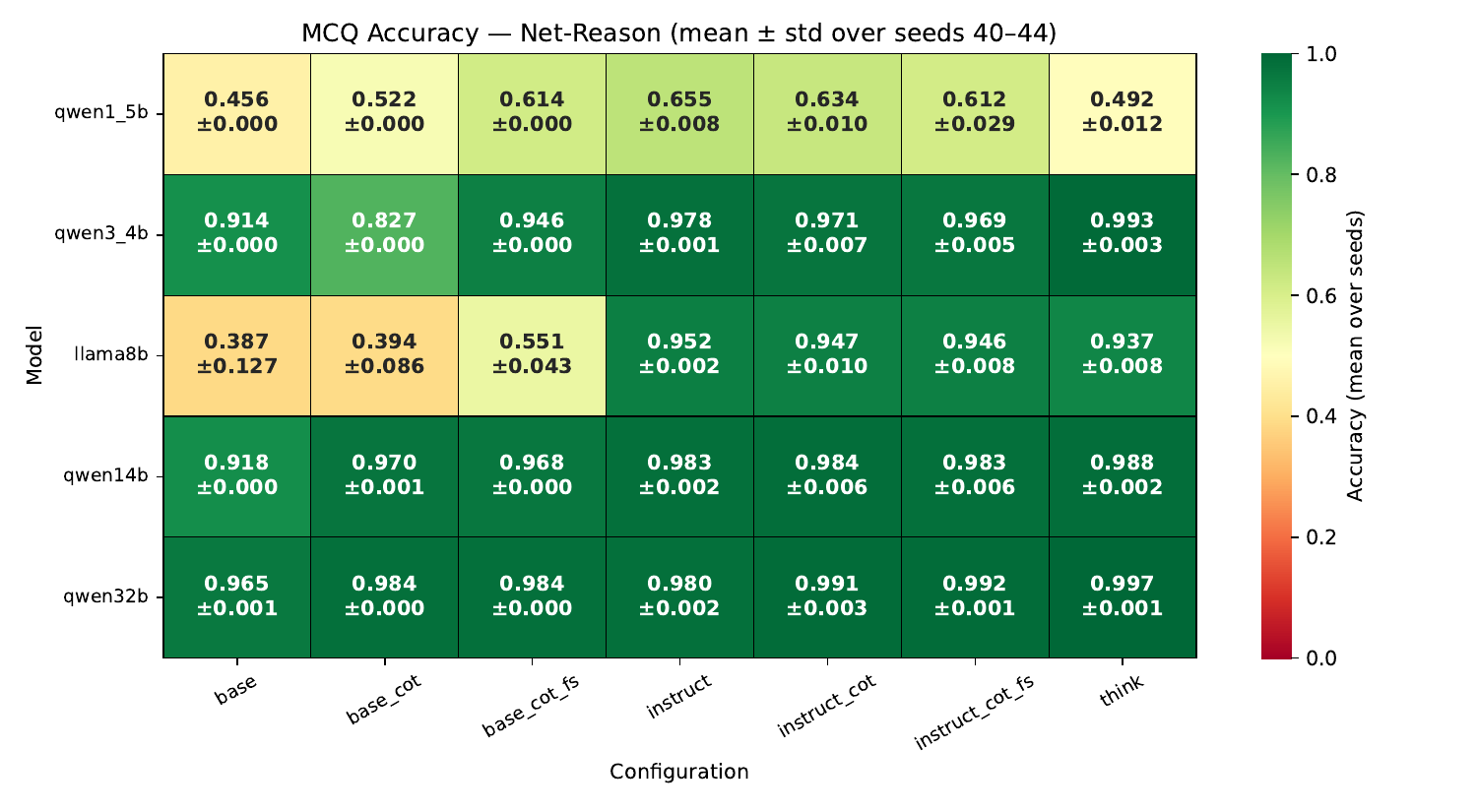}
        \caption{MCQ score}
        \label{fig:heatmap-overall-mcq}
    \end{subfigure}
    \hfill
    \begin{subfigure}{0.22\linewidth}
        \includegraphics[width=1\linewidth]{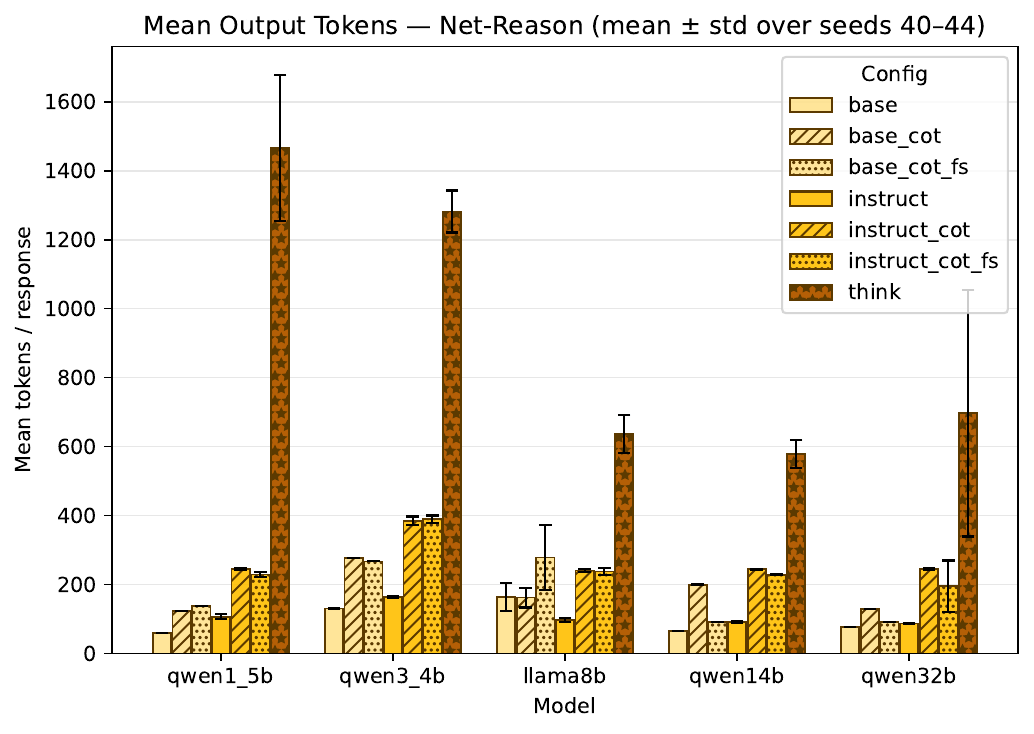}
        \caption{Overall Token Count}
        \label{fig:overall-token-count}
    \end{subfigure}
    \caption{Performance of selected models on Net-Reason - Average results over 5 seeds ($40$ to $44$). Figure \ref{fig:heatmap-overall} shows the accuracy of the selected models under different configurations. We observe that CoT prompting and reasoning versions of models generally outperform base or instruct counterparts. Figure \ref{fig:heatmap-overall-no-table} presents the accuracy of models when the Table is removed from the model's context. Notably, the performance drops, showing that the models actively rely on the Table to solve the questions. Figure \ref{fig:heatmap-overall-mcq} presents the inference of models when prompted to select an answer among four candidate answers. 
    Figure \ref{fig:overall-token-count} present the average token-count of the selected models on Net-Reason. We note that reasoning variant configurations are more costly, producing much more tokens overall.}
    \label{fig:results-overall-all}
\end{figure*}

\noindent \textbf{Question Types.} Net-Reason contains four question types that reflect progressively demanding reasoning over structured tables, which we summarized in 
the bottom row of Figure~\ref{fig:pipeline}.
\emph{Data Extraction} requires retrieving a single cell value, \emph{Aggregation} asks to combine multiple cells, \emph{Logic} requires combining knowledge to interpret a table, and \emph{Multi row/col} requires comparing multiple entries from the table. 

\noindent \textbf{Domain specificity.} Domain specificity adds a further layer of difficulty. Answering a Logic question about 5G QoS parameters correctly requires more than table-navigation skill: the model must also know, for example, that a lower packet-delay budget indicates stricter latency requirements, or that Delay Critical GBR is reserved for industrial automation use cases. These conventions are not spelled out in every table row, and models that lack this grounding will struggle even when the relevant cells are trivially identifiable. This makes our tabular dataset a more demanding proxy for real-world telecommunications QA than domain-agnostic tabular benchmarks.

\noindent \textbf{Dataset quality.} To estimate the quality of the final benchmark, three domain experts (graduate and post-graduate researchers in communication networks) independently audited the same $48$ questions, sampled uniformly across the four question types ($12$ per type). For each item, they judged whether the question is clear, whether it is answerable from the table, and whether the reference answer is correct (see Table \ref{tab:human-val-dataset-quality}). Across annotators, the questions were rated clear in 96.5\% of cases, fully answerable in 91.7\%, and the reference was correct in 95.8\%.

\begin{table}[t]
    \centering
    \small
    \begin{tabular}{lcccc}
        \toprule
        & A1 & A2 & A3 & All \\
        \midrule
        Clear question (\%) & 95.8 & 97.9 & 95.8 & 96.5 \\
        Fully answerable (\%) & 95.8 & 89.6 & 89.6 & 91.7 \\
        Reference correct (\%) & 100 & 100 & 87.5 & 95.8 \\
        \bottomrule
    \end{tabular}
    \caption{Human evaluation (48 items $\times$ 3 annotators).}
    \label{tab:human-val-dataset-quality}
\end{table}

\section{Experimental Design}
\label{sec:experimental-design}

We use Net-Reason to compare the accuracy and token cost of base, instruct and reasoning configurations, as well as agentic pipelines. We also use Net-Reason to test whether early-stopping methods reduce the cost of reasoning models. 

\begin{figure*}[h]
    \centering
    \begin{subfigure}{0.32\linewidth}
        \includegraphics[width=1.0\linewidth]{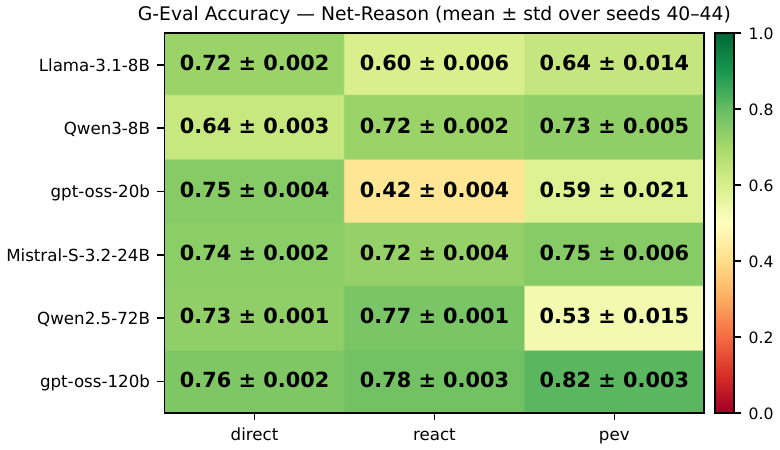}
        \caption{Overall accuracy}
        \label{fig:accuracy_agentic}
    \end{subfigure}
    \hfill
    \begin{subfigure}{0.32\linewidth}
        \includegraphics[width=1.0\linewidth]{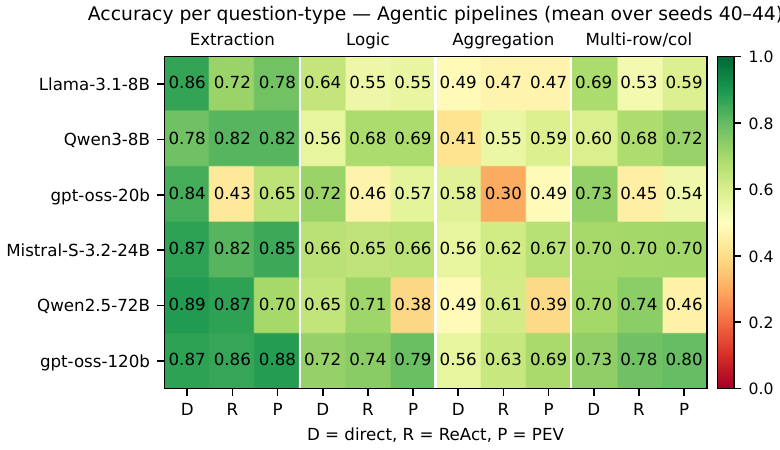}
        \caption{Accuracy per question-type}
        \label{fig:accuracy_per_type}
    \end{subfigure}
    \hfill
    \begin{subfigure}{0.32\linewidth}
        \includegraphics[width=1.0\linewidth]{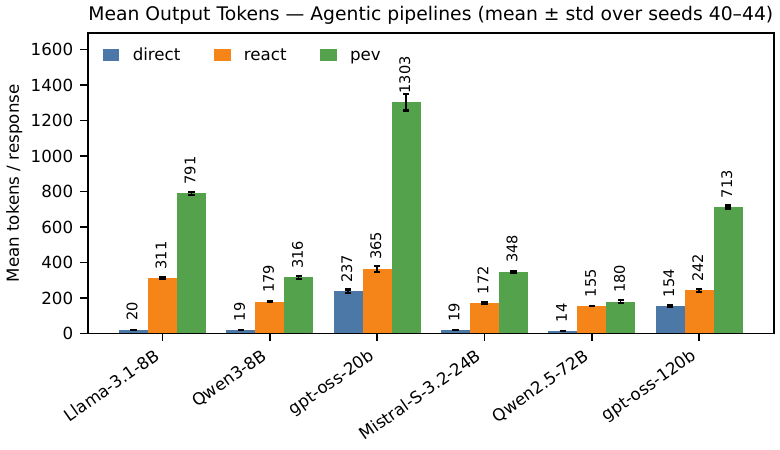}
        \caption{Overall Token Count}
        \label{fig:token_count_agentic}
    \end{subfigure}
    \caption{Performance of agentic pipelines (Direct, ReAct, PEV) on Net-Reason for six models, mean over seeds (40 to 44). Figure \ref{fig:accuracy_agentic} shows the overall performance of agentic pipelines on Net-Reason. Agentic pipelines helps selectively: ReAct and PEV improves performance over Direct for Qwen3-8B ($0.64$ vs. $0.73$), gpt-oss-120b ($0.76$ vs. $0.78$), and maginally Mistral-S-3.2-24B ($0.74$ vs. $0.75$). But the performance degrades llama-3.1-8B ($0.72$ vs. $0.60$) and gpt-oss-20b ($0.75$ vs. $0.42$), without clear dependance on model's size and family. Figure \ref{fig:accuracy_per_type} details the score per question-types. Gains from agentic pipelines are the highest on Aggregation and Muti-row/columns, where the tool-call usage seems the most natural, and are the smallest for Data Extraction (where Direct already performs well). Figure \ref{fig:token_count_agentic} presents the token-count per question. ReAct and PEV cost 1.5-16x and 4.6-40x more tokens than Direct, respectively.}
    \label{fig:agentic_pipeline_perf_netreason}
\end{figure*}

\noindent \textbf{Models.} We selected five open-weight models from 1.5 to 32B parameters (\texttt{Qwen2.5-1.5B}, \texttt{Qwen3-4B}, \texttt{Llama-3.1-8B}, \texttt{Qwen2.5-14B}, and \texttt{Qwen2.5-32B}) over 7 inference configurations, namely: Base, Base CoT, Base CoT Few-Shot (FS), Instruct, Instruct CoT, Instruct CoT FS, and Reasoning (see Table \ref{tab:selected-models}, Figure \ref{fig:cot_prompt} in Appendix C). Base, instruct and reasoning variants are different post-training releases of the same model family, so our comparisons are between available configurations of the same model architecture. For reproducibility, all models are inferred with greedy decoding, with five seeds per sample ($40$-$44$). 

\noindent \textbf{Agentic pipelines.} Further, we demonstrate that the Net-Reason benchmark can be used to evaluate agentic pipelines the same way we evaluated single call to models. We implemented three pipelines in LangGraph \cite{wang2024agentailanggraphmodular}, namely: \emph{Direct} (a single call, as our original inference pipeline), \emph{ReAct} \cite{yao2023reactsynergizingreasoningacting}, where the model alternates between writing Python scripts to analyze the table and reading the output, and \emph{Plan-Execute-Verify (PEV)}, which builds on ReAct, and adds a planning step before the tool-call \cite{wang2023planandsolvepromptingimprovingzeroshot}, and a verification step to get feedback \cite{madaan2023selfrefineiterativerefinementselffeedback}. Figure \ref{fig:agentic-pipelines} in Appendix E illustrates our different agentic pipelines. We evaluated six models (Llama3.1-8B-Instruct, Qwen3-8B, gpt-oss-20b, Mistral-Small-3.2-24B, Qwen-2.5-72B, and gpt-oss-120b) over five seeds ($40$ to $44$). 

\noindent \textbf{Metrics.} We used \emph{G-Eval} \cite{liu2023gevalnlgevaluationusing} using \texttt{GPT-4o-mini} as an LLM-as-a-Judge. The model gives a score between 0 and 1 to assess the correctness of the answer based on the ground-truth. Figure \ref{fig:g_eval_prompt} in Appendix D presents the prompt.

\begin{figure}[t]
    \centering
    \includegraphics[width=0.9\linewidth]{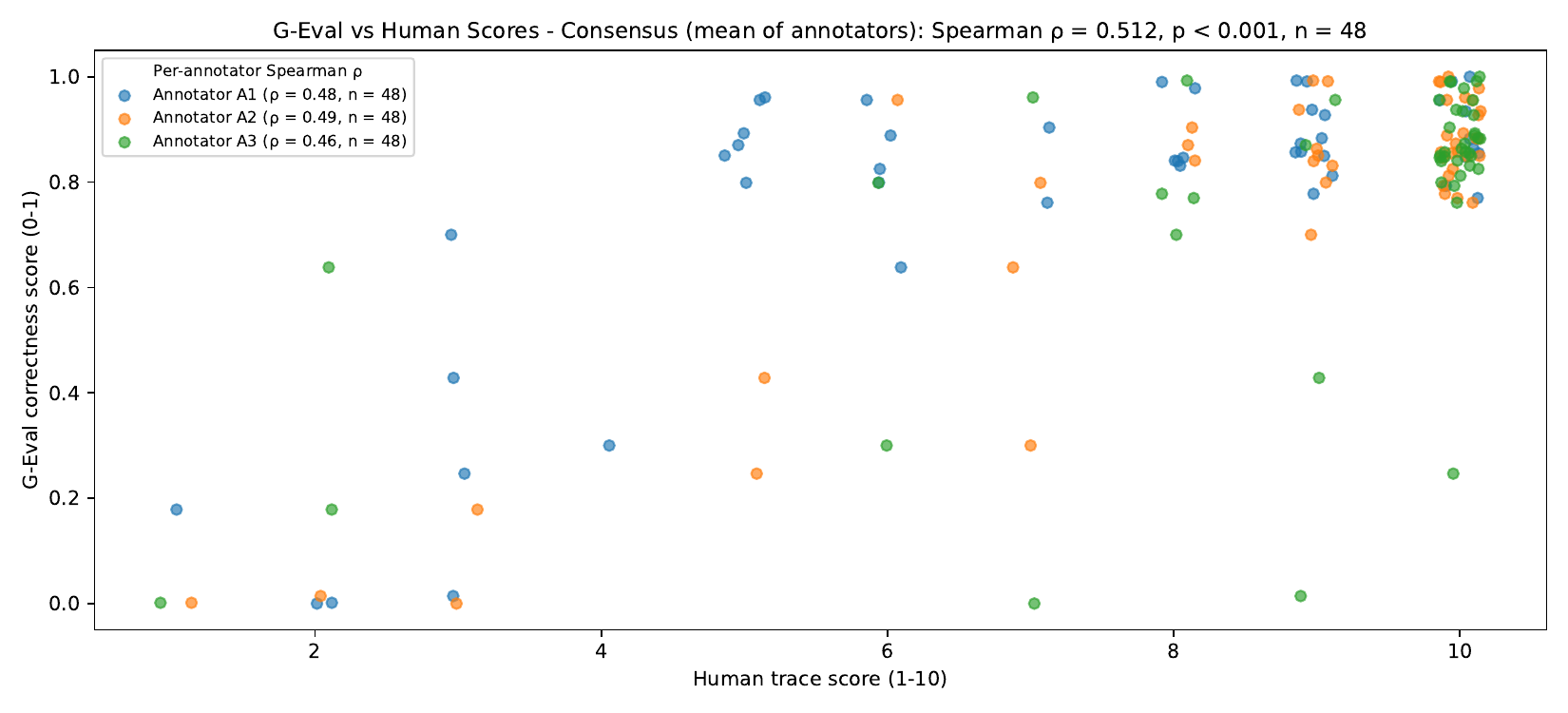}
    \caption{Human validation}
    \label{fig:g-eval-vs-human}
\end{figure}

\noindent \textbf{Validation of the scoring metric.} While G-Eval is widely adopted by the literature, we further validate its applicability as a scoring metric for Net-Reason. The same three annotators also rated, on a scale from $1$ to $10$ and with the G-Eval rubric (see Figure \ref{fig:g_eval_prompt} in Appendix D), the answer from DS-Llama-8B to the same $48$ questions. Figure \ref{fig:g-eval-vs-human} presents the Spearman correlation between the G-Eval score and the human rating. We obtained a correlation score of $\rho=0.512$ (statistically significant), showing a moderate agreement. The annotators agree well when the model is correct and incorrect (see clusters on top right and bottom left).
We therefore treat G-Eval as a satisfying proxy for the model's accuracy.

\noindent \textbf{Efficient test time-scaling methods.} Reasoning models can be inefficient \cite{liu2025efficientinferencelargereasoning} and inference-time
efficiency methods are effective in limiting the verbosity of LRMs. We included three types of methods in our initial analysis: 
(1) Prompting methods, where we instruct models to not be verbose \cite{lee2025llmscompresschainofthoughttoken,han2025tokenbudgetawarellmreasoning}. (2) TRACES, a black-box method which early-stops the text generation based on the type of reasoning steps \cite{belkhiter2026tracestaggingreasoningsteps}. (3) EAT \cite{wang2026entropythinkreasoningmodel} and DEER \cite{yang2025dynamicearlyexitreasoning}, gray-box and black-box methods, respectively, which early-exit the generation based on the model's confidence. To reproduce the algorithms, we used hyperparameter $\delta = 0.9$ for TRACES, $\lambda = 0.95$ for DEER, and ($\alpha=0.2$, $\delta=10^{-3}$) for EAT. Inference parameters are listed in Appendix C.

\section{Evaluation}

In this section, we first study the performance and costs of models across configurations, and agentic pipelines. Further, we compare the performance per question-type, and we analyze the performance of adaptive reasoning methods.

\noindent \textbf{Reasoning variants outperform base models.}
Figure \ref{fig:results-overall-all} presents the results of the different configurations. Overall, we observe that reasoning variants (CoT prompting or training) systematically outperform base or simple instruct models. This finding supports that our Net-Reason dataset contains questions that require some form of reasoning. 

Specifically, for base and instruct models, we observe that CoT prompting almost systematically matches or improves the model's performance (up to +16 points). Importantly, reasoning models (think) improve performance from 13 to 42 points compared to base models. Furthermore, newer model's version matter as much as the model's size ($0.841$ for Qwen3-4B Think vs. $0.842$ for Qwen2.5-32B Think).

\begin{figure*}[t]
    \centering
    \begin{subfigure}{0.32\linewidth}
        \includegraphics[width=1\linewidth]{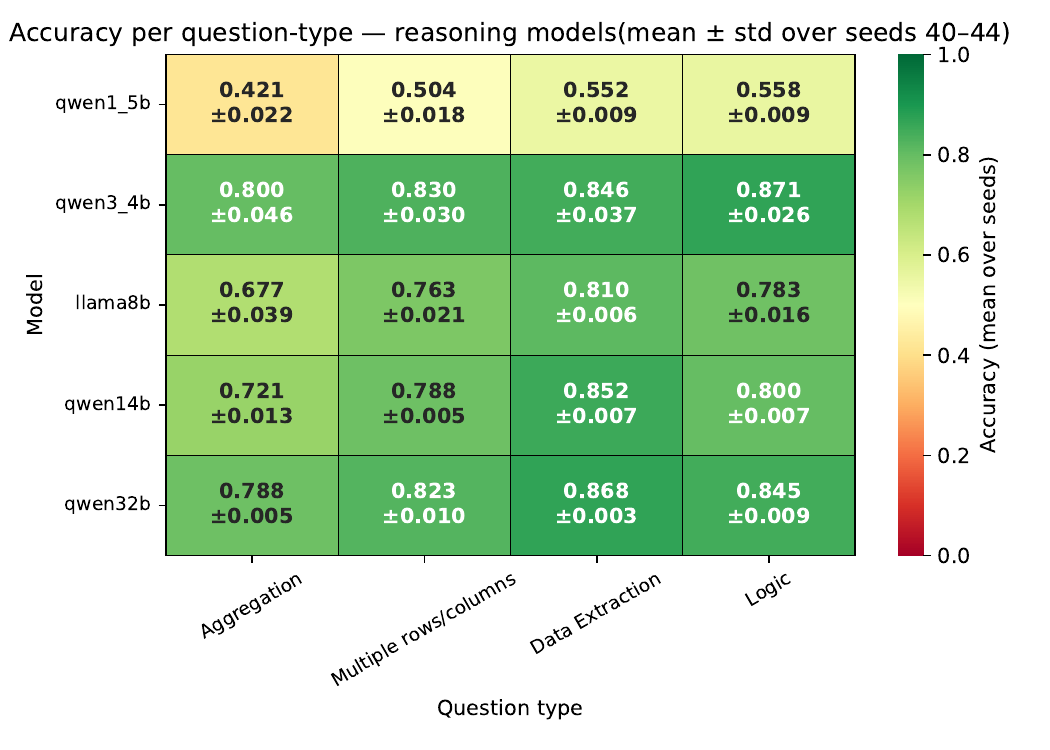}
        \caption{Accuracy per question-type}
        \label{fig:score-think-per-question}
    \end{subfigure}
    \hfill
    \begin{subfigure}{0.32\linewidth}
        \includegraphics[width=1\linewidth]{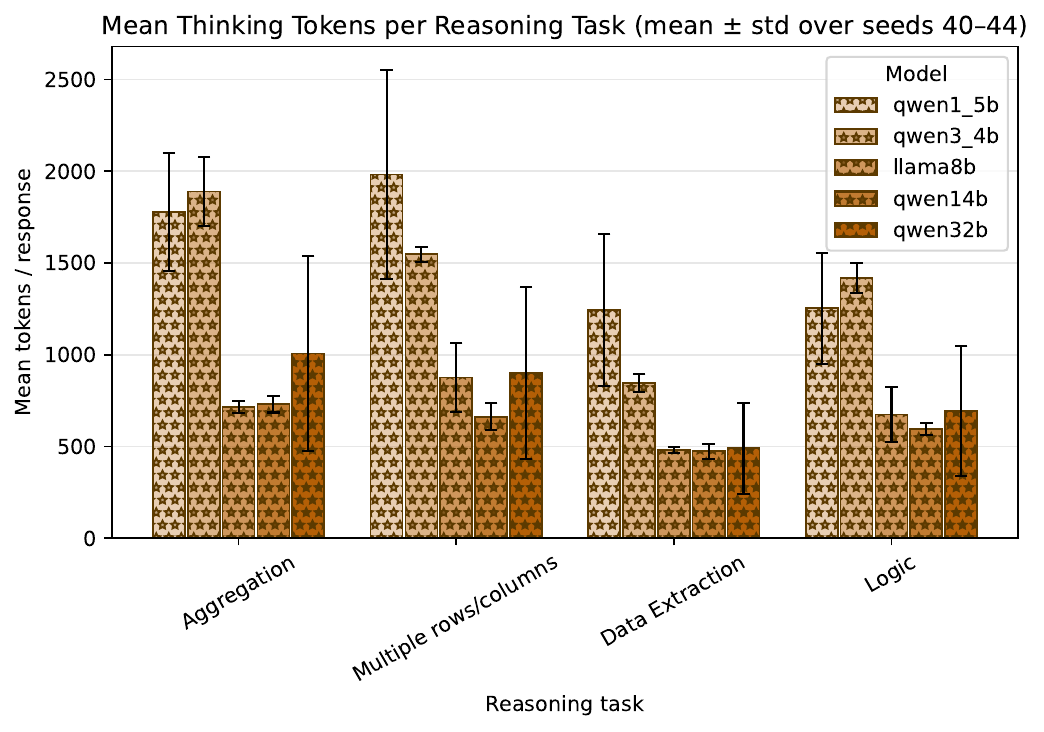}
        \caption{Token-count per question-type}
        \label{fig:token-think-per-question}
    \end{subfigure}
    \hfill
    \begin{subfigure}{0.32\linewidth}
        \includegraphics[width=1\linewidth]{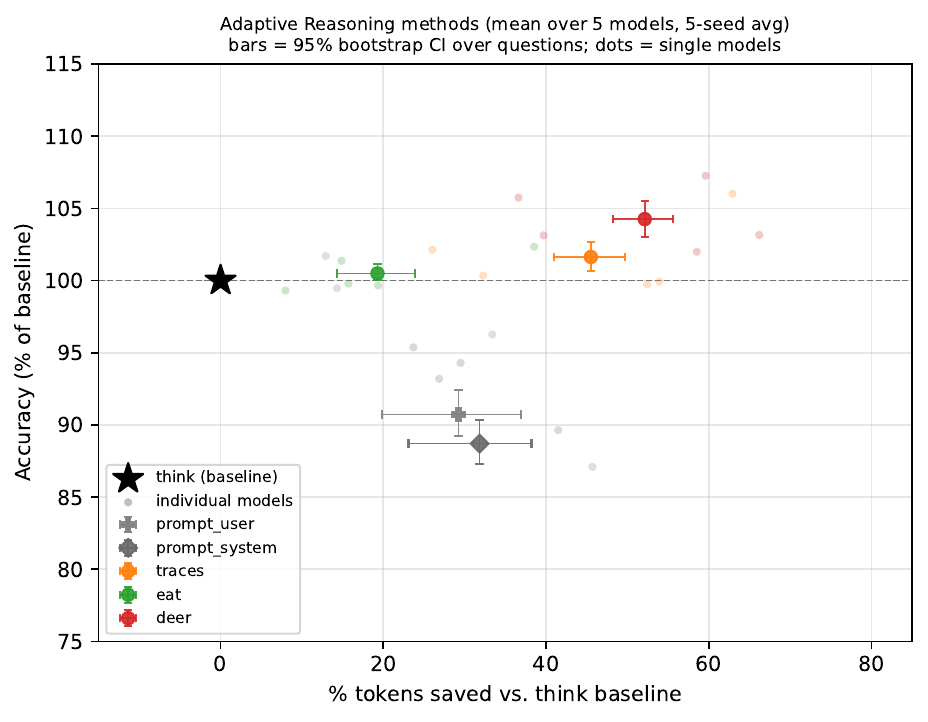}
        \caption{Early-Exit - Overall}
        \label{fig:early-exit-overall}
    \end{subfigure}
    \\
    \begin{subfigure}{0.24\linewidth}
        \includegraphics[width=1\linewidth]{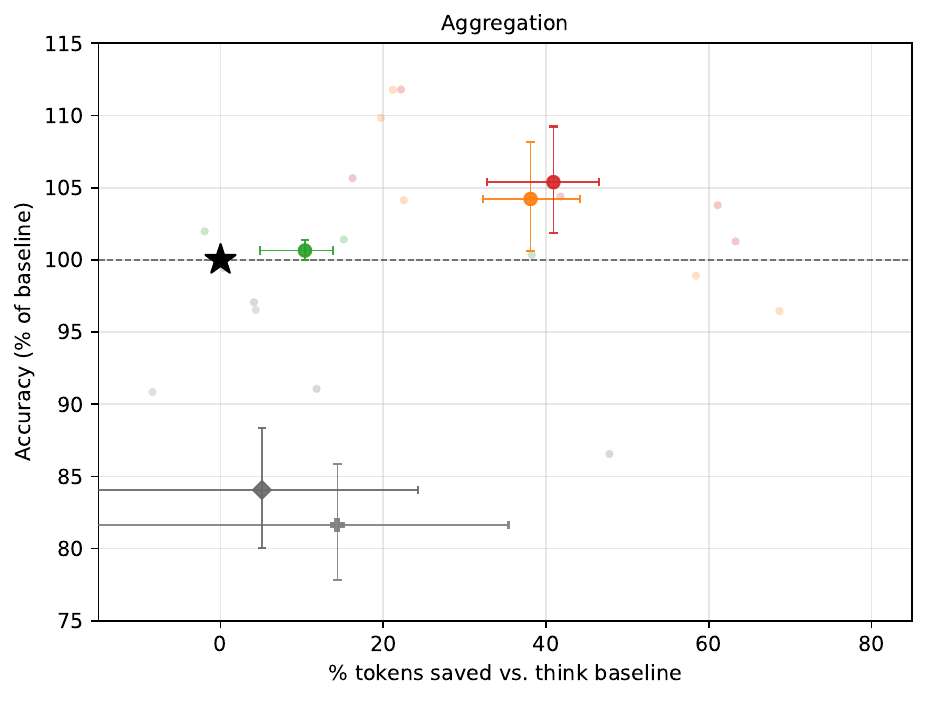}
        \caption{Early-Exit - Aggregation}
        \label{fig:early-exit-aggregation}
    \end{subfigure}
    \hfill
    \begin{subfigure}{0.24\linewidth}
        \includegraphics[width=1\linewidth]{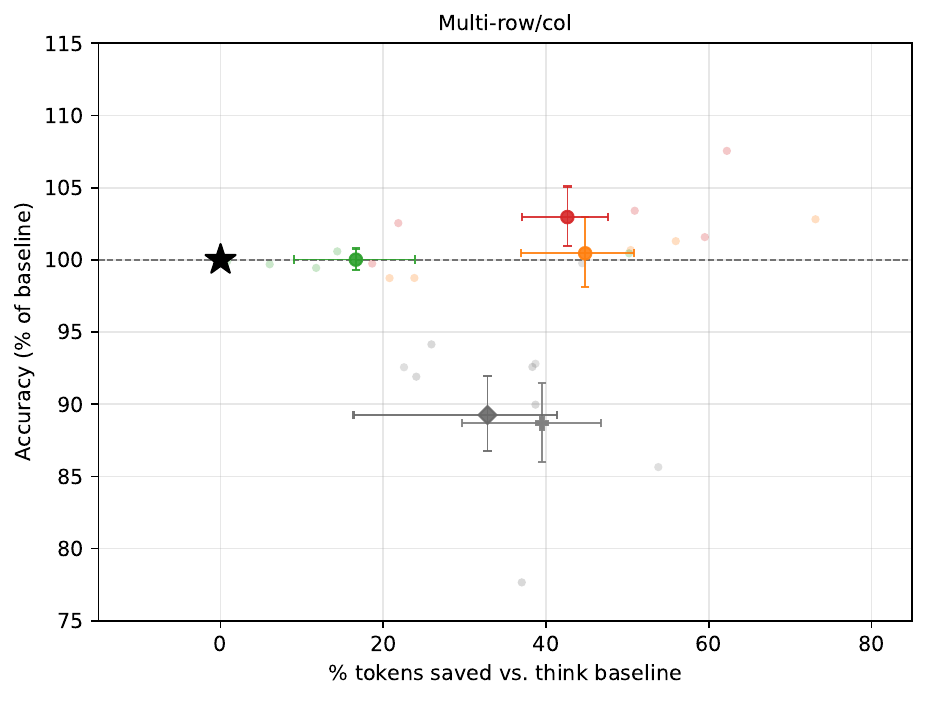}
        \caption{Early-Exit - Multi rows/col.}
        \label{fig:early-exit-multi}
    \end{subfigure}
    \hfill
    \begin{subfigure}{0.24\linewidth}
        \includegraphics[width=1\linewidth]{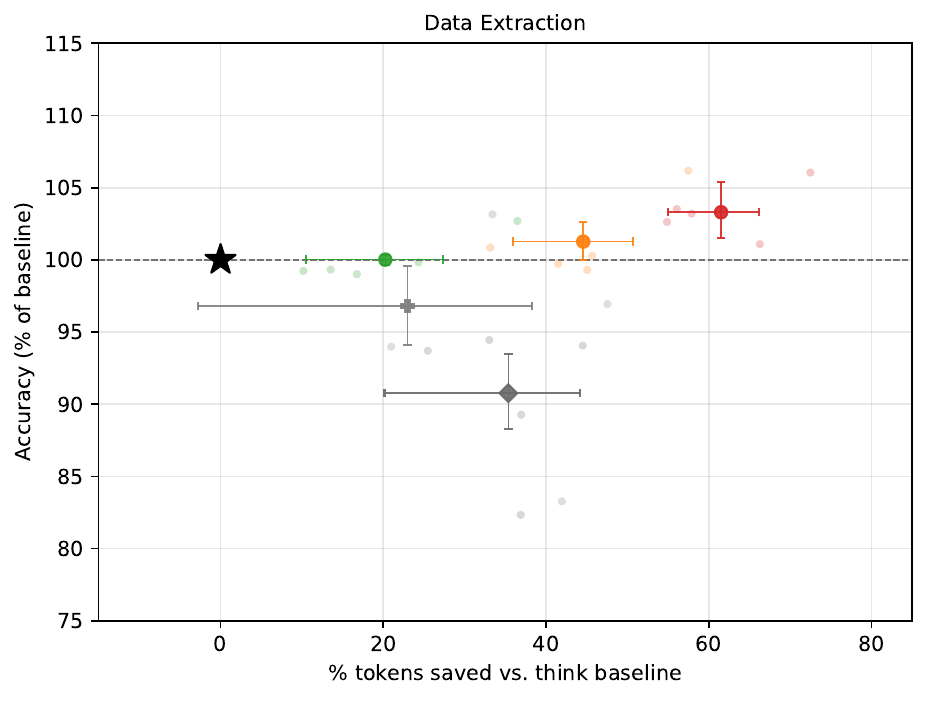}
        \caption{Early-Exit - Data Extraction}
        \label{fig:early-exit-extraction}
    \end{subfigure}
    \hfill
    \begin{subfigure}{0.24\linewidth}
        \includegraphics[width=1\linewidth]{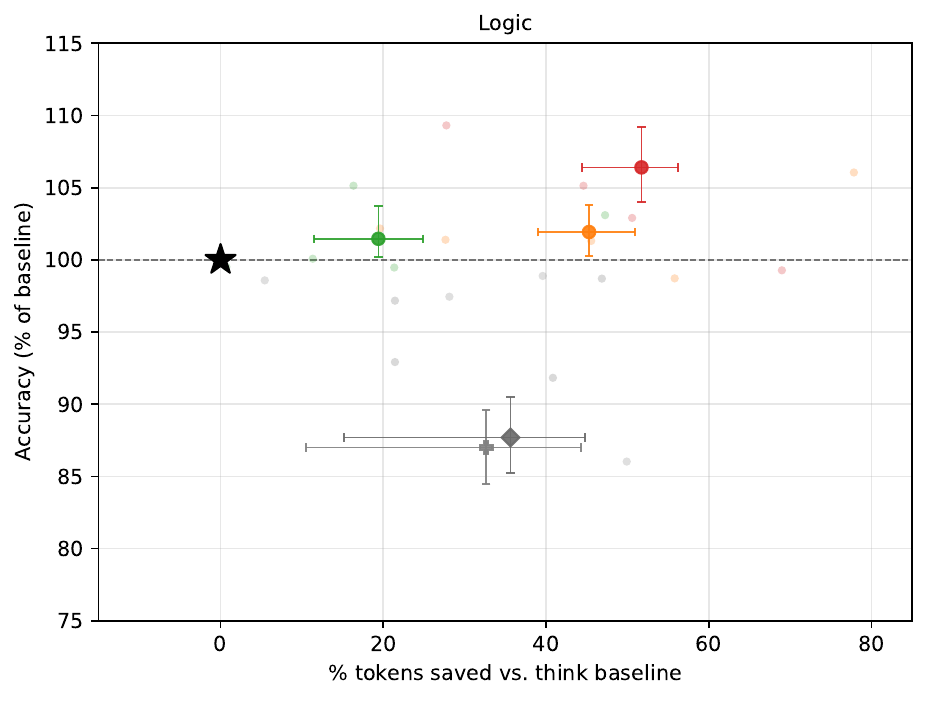}
        \caption{Early-Exit - Logic}
        \label{fig:early-exit-logic}
    \end{subfigure}
    \caption{Accuracy/Token-count trade-off of selected reasoning models on Net-Reason, per question-type, over five seeds ($40$ to $44$). Figures \ref{fig:score-think-per-question} and \ref{fig:token-think-per-question} show the accuracy and token-count, respectively, of the reasoning models per question-type. We note that \emph{Aggregation} and \emph{Multiple rows/columns} questions are the most challenging, resulting in lower accuracy and higher token-count (from 1.5x to 2.5x the token-count of the two other question types). Figure \ref{fig:early-exit-overall} presents the overall savings of early-exit methods, averaged across all models. We demonstrate the applicability of these methods to the dataset, with token-count saving ranging from 10 to 50\% and retained accuracy between 87 to 105\% compared to standard inference. Further, Figures \ref{fig:early-exit-aggregation}-\ref{fig:early-exit-logic} compare the performance of these methods per question-type. The performance of adaptive reasoning methods depends on the question-type.} 
    \label{fig:results-per-question}
\end{figure*}

\noindent \textbf{Table dependencies.} Figure \ref{fig:heatmap-overall-no-table} presents the performance of models when removing the Table from their context. We observe a relative drop from -45\% to -67\% accuracy across all models and configurations. It shows that model cannot answer most of the Net-Reason benchmark without a Table. Yet, the performance is not null ($0.15$-$0.38$), meaning that some questions can still be answered partially without the Table. This is because some documents only supply general knowledge, and some questions can be answered with the model's trained or prior knowledge itself. Further, G-Eval's rubric awards partial credits, resulting in non-null performance.

\noindent \textbf{MCQ Net-Reason.} Further, what makes the evaluation difficult is that the answer is free-form. While mathematical benchmarks can use a library to parse the ground-truth answer, knowledge and reasoning benchmark often use MCQ prompting to evaluate the performance of models. For this reason, we replicate this evaluation framework and for each sample, we prompted the model with four potential answers. Figure \ref{fig:heatmap-overall-mcq} presents the performance of models with MCQ-prompting. The MCQ accuracy is near ceiling ($0.94$-$1.00$) for most instruct and reasoning configurations of $\geq$4B, so MCQ separates their performance poorly. The gap with the G-Eval free-form scores (e.g. $0.599$ vs. $0.983$ for Qwen-2.5-14B Instruct) suggests that the MCQ task is easier because it narrows down the number of possibilities for the model to answer. Importantly, we observe that the ordering performance of models is largely preserved, with exception of Qwen-2.5-1.5B and Llama-3.1-8B where reasoning variants scores at or slightly bellow instruct models.

\noindent \textbf{Cost of reasoning.} While reasoning variants outperform base models, reasoning models are much more costly. Indeed, Figure \ref{fig:overall-token-count} shows the overall token-count per variant across models. We systematically observe that CoT prompting increases the token-count of base models, while reasoning models cost 2 to 15 times more tokens than the respective base versions. In Figure \ref{fig:results-per-question}, we will study this further.

\noindent \textbf{Agentic pipelines.} Figure \ref{fig:agentic_pipeline_perf_netreason} shows the performance of our agentic pipelines on Net-Reason. Our findings suggest that agentic pipelines did not uniformly help, and depends on the LLM selected. \emph{ReAct} and \emph{PEV} improve Qwen3-8B (up to $0.73$ vs. $0.64$ using \emph{Direct}) and gpt-oss-120b (up to $0.82$ vs. $0.76$). Results are mixed for Qwen2.5-72B (improves only for \emph{ReAct}, $0.77$ vs. $0.73$) and Mistral-S-3.2-24B (improves only for \emph{PEV}, $0.75$ vs. $0.74$). And \emph{ReAct} and \emph{PEV} lowers performance on Llama-3.1-8B and gpt-oss-20b. The agentic pipelines also increases token-count from 1.5–16× (ReAct) and 4.6–40× (PEV) times compared to \emph{Direct} LLM call. Importantly, the performance and costs of agentic pipelines also depends on the question-type, with higher accuracy obtained on Data Extraction across agentic pipelines.

\noindent \textbf{Performance by question type.} Next, Figures \ref{fig:score-think-per-question} and \ref{fig:token-think-per-question} present the performance and token-count per question-type across models, respectively. First, we note that models obtained the highest performance on the \emph{Data Extraction} and \emph{Logic} categories. Conversely, we observe that \emph{Aggregation} and \emph{Multi rows/col.} questions systematically produces more tokens compared to the other two categories (around 1.5-2.5x more), with lower accuracy. These findings show that the question types have different efficiency trade-off. 

\noindent \textbf{Adaptive Reasoning methods.} Further, we applied five adaptive reasoning methods to Net-Reason across selected reasoning models, namely: prompting at the user and system level, TRACES, EAT and DEER. Prompting methods achieve 20 to 40\% token-count saving, but at the cost of the accuracy (up to -13 point accuracy drop). Prompting methods are also less reliable, with larger CIs in both accuracy savings, and saved token-count. In comparison, TRACES and DEER achieve 40 to 50\% token-count reduction, while matching and improving the accuracy of standard inference. 

\noindent \textbf{Early stopping by question type.} Further, we compare the efficiency of early-exit methods per question-type. Interestingly, the saving depend on the question-type. Methods applied to \emph{Aggregation} and \emph{Multi rows/cols} offers 20 to 40\% token-count saving, while \emph{Data Extraction} and \emph{Logic} offers up to 40\% to 60\% token-count saving. 

\section{Conclusion and future work}

We introduce Net-Reason, a novel in-domain Table Question-Answer benchmark for Communication Networks. Net-Reason contains four question types to evaluate the ability of models to reason over various types of technical documents. We evaluated 5 LLMs across 7 inference configurations, and found that reasoning version systematically outperforms base counterparts. This finding supports that Net-Reason requires some form of reasoning, and demonstrate the value of reasoning models in Network environments. Our evaluation of agentic pipelines further shows that external orchestration can improve correctness, but its benefits depend on the underlying model and question category, despite substantially increased token expenditure. Moreover, we showed that early-stopping methods are effective in this domain, and should be further explored.

\noindent \textbf{Future work.} While these results highlight the potential of LRMs on tasks applied to Networks, future work will focus on expanding our dataset. For instance, creating questions over multi-tables could increase the complexity of the questions. Assessing robustness by perturbing tables could also add value. Further, the Net-Reason benchmark is an opportunity to create temporal benchmark, where the model should reason about Network configurations across time.

\bibliography{aaai2027}


\newpage
\appendix
\onecolumn

\section*{Appendix}

\subsection*{LLM Usage}

We acknowledge the use of Large Language Models for the purpose of our experimentation in our paper (for running our dataset generation pipeline, as well as for benchmarking models on NetReason). Specifically, as stated in the Section \emph{Net-Reason dataset}, the question-answer pairs in this dataset were generated with the assistance of a large language model (\texttt{Opus-4.5} \footnote{see \url{https://www.anthropic.com/news/claude-opus-4-5}}). Following automated extraction of text fragments and tabular content from the source documents, Claude was used to generate candidate question-answer pairs grounded in those extracted passages. The model generated questions requiring reasoning over structured markdown tables, covering data extraction, aggregation, and multi-row logic. 

All model-generated pairs subsequently underwent human validation to verify that each question is well-formed, unambiguous, and appropriately answered by the cited passage or table. Questions that did not meet these criteria were revised or discarded. The source documents themselves, the text extraction pipeline, and the final validation judgements were produced entirely by the human authors. As well, as stated in the Section \emph{Experimental Design}, we relied on \emph{G-Eval} \cite{liu2023gevalnlgevaluationusing} using \texttt{GPT-4o-mini} \cite{openai2024gpt4ocard} to evaluate the inferences of models.

\subsection{Limitations}

We use output-token count as a proxy for inference cost, which does not capture differences in model size, input-token processing, tool execution, or inference latency. Our evaluation is restricted to selected open-weight models and does not include frontier closed-weight models as benchmarked systems. Furthermore, our validation of G-Eval is limited to 48 samples from a single reasoning model and indicates a moderate correlation with human ratings. Further validation across models and alternative scoring rubrics would strengthen the reliability of the evaluation. Finally, Net-Reason evaluates reasoning over supplied technical evidence rather than end-to-end network operations.

\newpage

\subsection{A. Source Documents}

The dataset is built from 36 technical documents drawn from the telecommunications standards ecosystem, selected to cover the breadth of 5G Quality-of-Service (QoS) concepts. The corpus spans multiple standards bodies, file formats, and levels of technical abstraction, which we describe below.

\paragraph{3GPP specifications.}
The backbone of the corpus consists of formal 3GPP technical documents like~\cite{3gpp_ts23501, 3gpp_ts22261}, covering 5G System architecture, the standardised 5G QoS Identifier (5QI) table, and service requirements for Ultra-Reliable Low-Latency Communication (URLLC) scenarios. These documents combine dense technical prose with multi-column parameter tables, where correct answers often require cross-referencing several fields simultaneously: resource type, priority level, packet-delay budget, packet-error rate, and example service class.

\paragraph{IETF and ITU-T standards.}
The corpus also includes IETF and ITU-T standards like~\cite{ash2010rfc5976, itu_y1541_2011, contreras2024dscp}, which introduce the IP-layer QoS class framework that complements 3GPP's radio-layer definitions. Written in tightly regulated standards prose, these documents contain structured tables mapping QoS classes to network-layer parameters, adding cross-layer reasoning challenges
that 3GPP specifications alone do not provide.

\paragraph{Industry white papers and technology reviews.}
The corpus also includes industry white papers and technology reviews like~\cite{5gamericas2019urllc, ericsson2018ltem, alriksson2021xr5g, cisco2001voipqos, muralidharan2017videoqos}, covering URLLC, LTE-M/NB-IoT, XR over 5G, and VoIP and video QoS design. These documents sit between formal standards and practical engineering guides, describing deployment trade-offs and configuration decisions in a style that requires domain knowledge to interpret. 

\paragraph{Research papers.}
The corpus also draws from academic literature documents like~\cite{popovski2018slicing, khan2023cv2x, kosmanos2022v2x, gapeyenko2022xr5g, hazarika2023arvr, matz2020nbiot, haghani2010trafficaware, xavier2024openran, allbesmart2023oaibox}, covering network slicing across eMBB, URLLC, and mMTC, C-V2X for autonomous driving, XR standardisation, ultra-low-latency AR/VR, NB-IoT QoS, adaptive video streaming, Open RAN measurements, and a practical open-source 5G deployment guide. These sources introduce probabilistic and systems-level reasoning that goes beyond simple table look-up and that is absent from most telecom QA corpora.

\paragraph{Online reference portals.}
Five widely used reference portals provide concise, table-heavy summaries of 5G QoS parameters: the 5G-Tools QoS reference~\cite{5gtools2019qos}, the Award Solutions QoS evolution guide~\cite{awardsolutions_qosevolution}, the RF Wireless World 5QI table~\cite{rfwirelessworld_5qi}, the ShareTechnote 5QI reference~\cite{sharetechnote_5qi}, and the Techplayon 5G NR QoS overview~\cite{techplayon_qos}. These pages are dense in structured tabular content and serve as a natural source of Data Extraction and Aggregation questions in the tabular subset.

\paragraph{Configuration references.}
The srsRAN gNB YAML configuration reference~\cite{srsran_config} provides structured, semi-formal technical content and details about how to configure a 5G base station and what configuration parameters are available.

\medskip
Taken together, the 36 documents span a vertical slice of the 5G knowledge stack: from chip-level timing parameters (srsRAN) to network-architecture principles (3GPP TS 23.501) to end-to-end service requirements (AR/VR latency, V2X reliability). This breadth requires models to reason across abstraction levels rather than exploiting vocabulary patterns from a single source. The coexistence of formal standards and practitioner guides also creates genuine terminological variation: the same concept can appear under different names and in different units across documents, which is a non-trivial disambiguation challenge for both extractive and tabular tasks.

\newpage

\subsection{B. Analysis of Net-Reason} 

\begin{figure}[h]
    \centering
    \begin{subfigure}{0.4\linewidth}
        \includegraphics[width=1\linewidth]{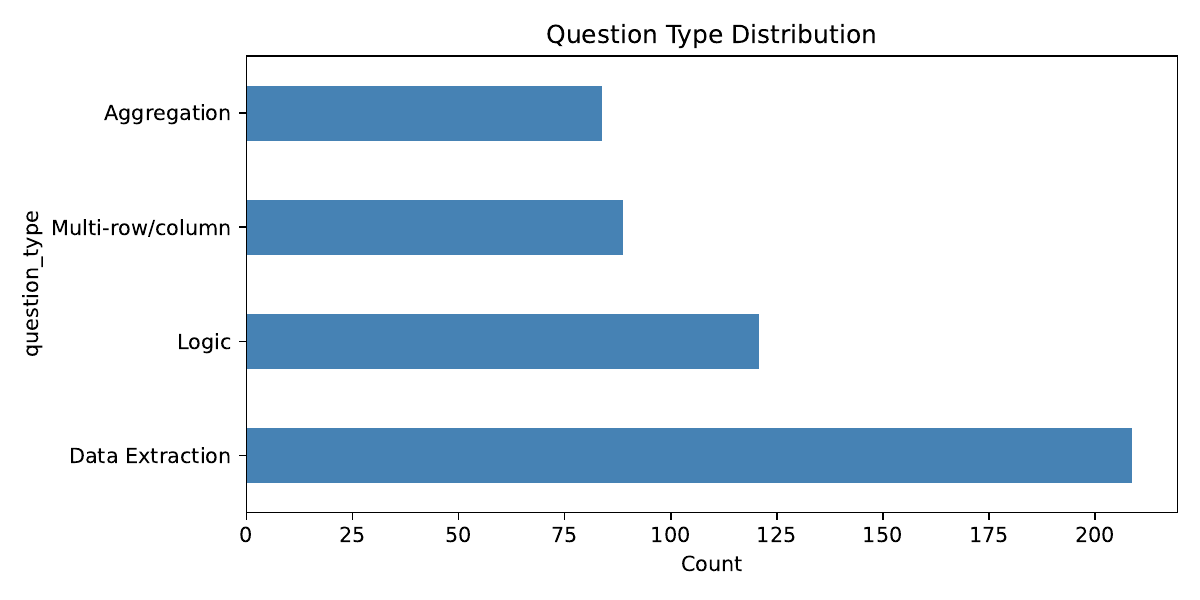}
        \caption{Distribution of Question-Type}
        \label{fig:distrib-question-type}
    \end{subfigure}
    \hfill
    \begin{subfigure}{0.55\linewidth}
        \includegraphics[width=1\linewidth]{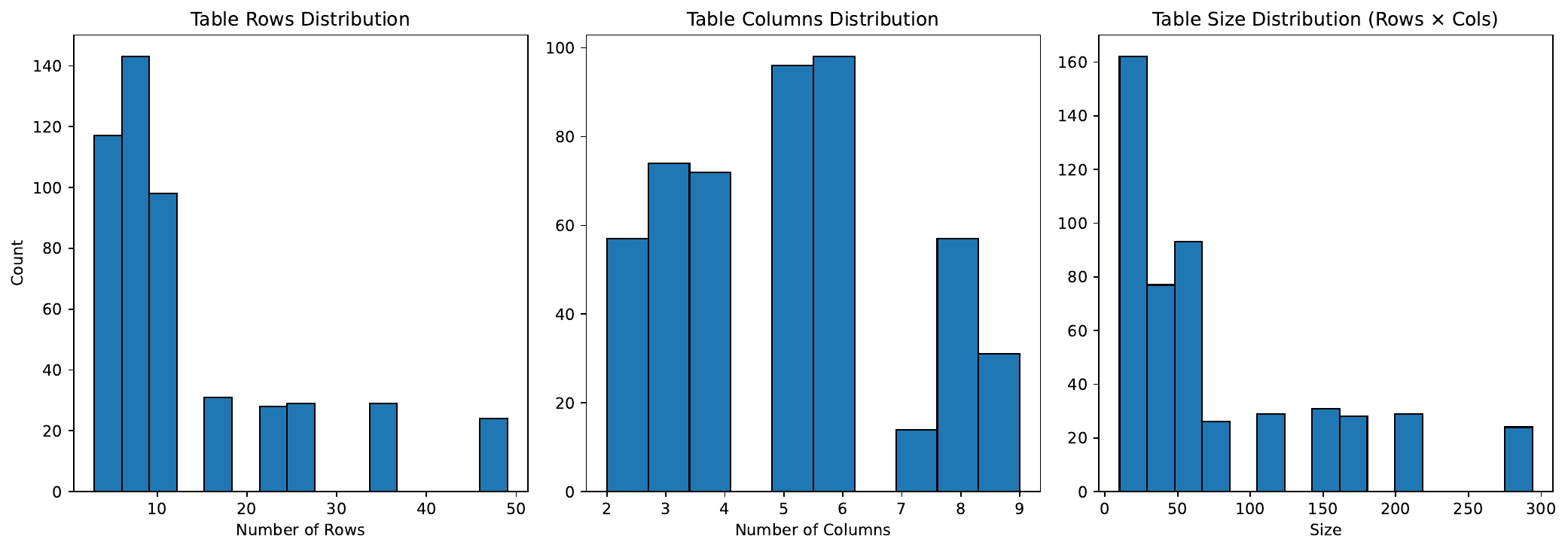}
        \caption{Size of Tables}
        \label{fig:opening-words}
    \end{subfigure}
    \begin{subfigure}{0.45\linewidth}
        \includegraphics[width=1\linewidth]{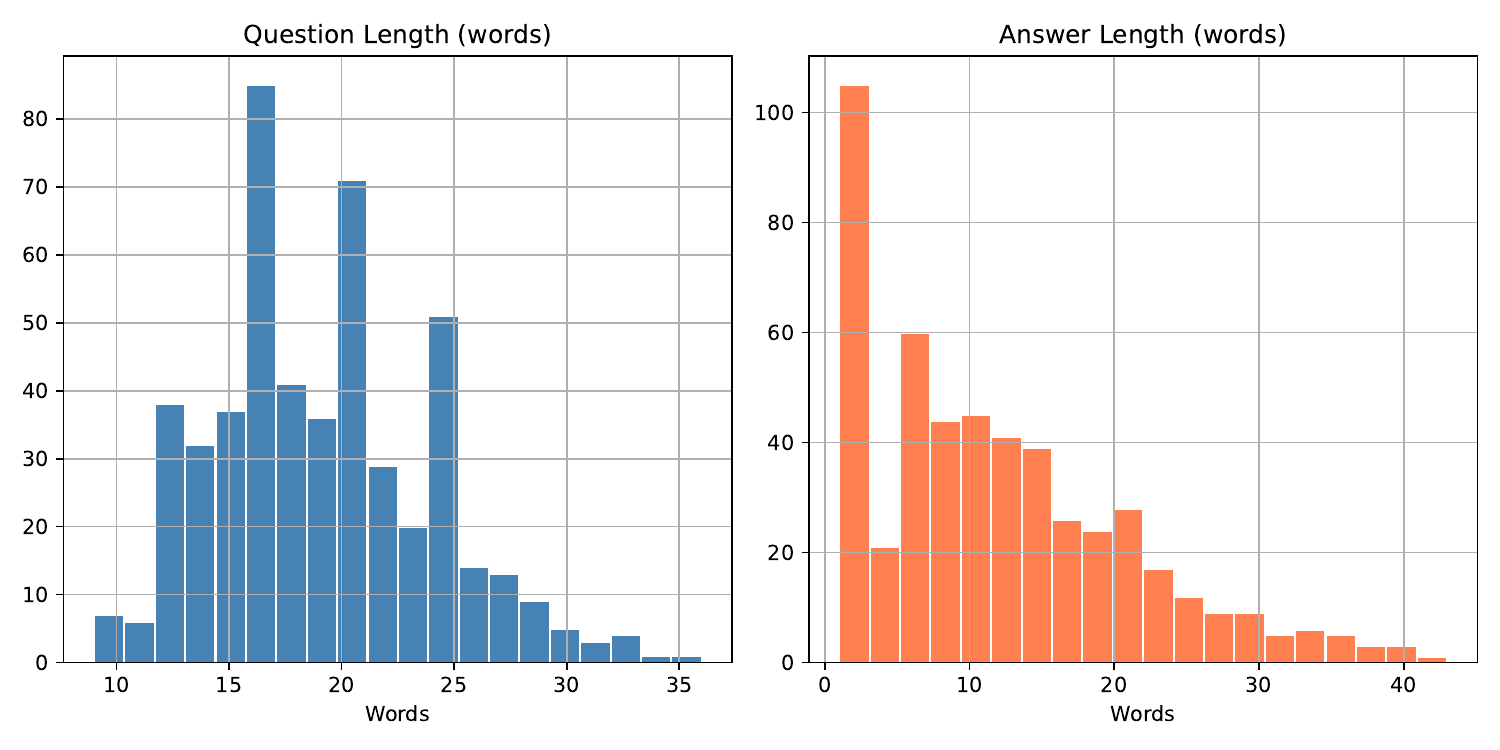}
        \caption{QA length distribution}
        \label{fig:qa-length-distrib}
    \end{subfigure}
    \hfill
    \begin{subfigure}{0.45\linewidth}
        \includegraphics[width=1\linewidth]{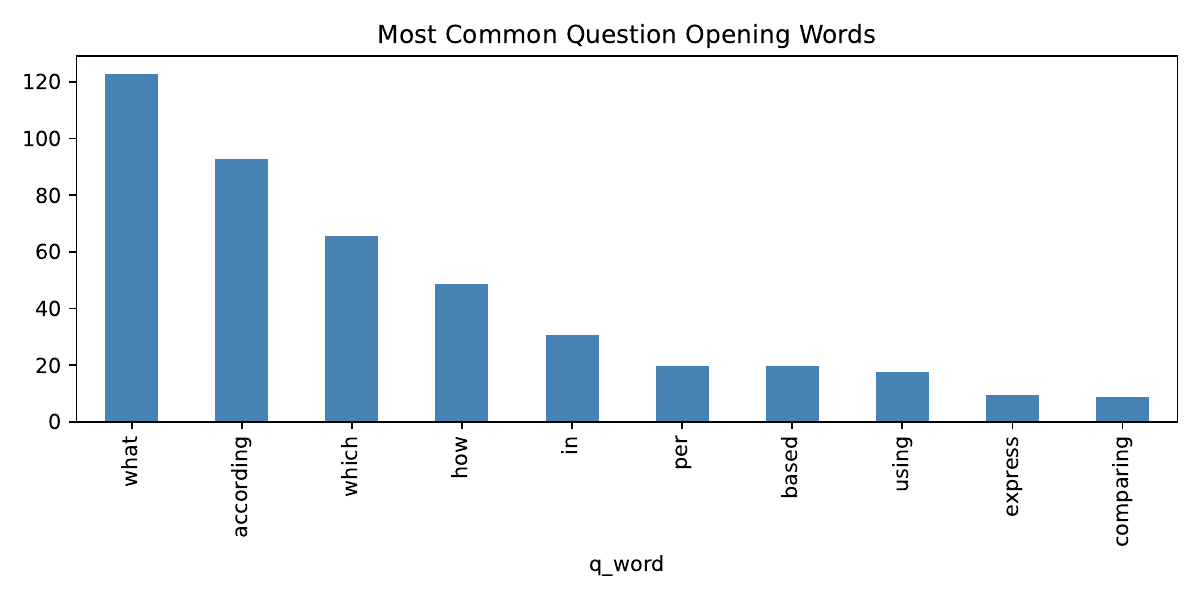}
        \caption{Question's opening words}
        \label{fig:answer-length}
    \end{subfigure}
    \caption{Analysis of Net-Reason}
    \label{fig:net-reason-analysis}
\end{figure}

Figure \ref{fig:net-reason-analysis} presents a set of statistic analysis of Net-Reason. Importantly, we show that each categories contains at least $84$ questions. Additionally, we show that the table used are diverse in term of size. The average number of rows is around $14$, the average number of columns is around $5$, and the average number of cells per table is around $77$.

\newpage

\subsection{C. Selected models and CoT prompting}

\begin{table}[h]
    \centering
    \tiny
    \begin{tabular}{ccccc}
        \toprule
        \textbf{Model type} & \textbf{Ref.} & \textbf{Base} & \textbf{Instruct} & \textbf{Reasoning} \\
        \midrule
        \texttt{Qwen2.5-1.5B} & \cite{qwen2.5} & \texttt{Qwen/Qwen2.5-1.5B} & \texttt{Qwen/Qwen2.5-1.5B-Instruct} & \texttt{deepseek-ai/DeepSeek-R1-Distill-Qwen-1.5B} \\
        \midrule
        \texttt{Qwen3-4B} & \cite{qwen3technicalreport} & \texttt{Qwen/Qwen3-4B-Base} & \texttt{Qwen/Qwen3-4B-Instruct-2507} & \texttt{Qwen/Qwen3-4B-Thinking-2507} \\
        \midrule
        \texttt{Llama-3.1-8B} & \cite{grattafiori2024llama3herdmodels} & \texttt{meta-llama/Llama-3.1-8B} & \texttt{meta-llama/Llama-3.1-8B-Instruct} & \texttt{deepseek-ai/DeepSeek-R1-Distill-Llama-8B} \\
        \midrule
        \texttt{Qwen2.5-14B} & \cite{qwen2.5} & \texttt{Qwen/Qwen2.5-14B} & \texttt{Qwen/Qwen2.5-14B-Instruct} & \texttt{deepseek-ai/DeepSeek-R1-Distill-Qwen-14B} \\
        \midrule
        \texttt{Qwen2.5-32B} & \cite{qwen2.5} & \texttt{Qwen/Qwen2.5-32B} & \texttt{Qwen/Qwen2.5-32B-Instruct} & \texttt{Qwen/QwQ-32B} \\
        \bottomrule
    \end{tabular}
    \caption{Selected models}
    \label{tab:selected-models}
    \vspace{-0.4cm}
\end{table}

\begin{figure}[h]
\centering
\footnotesize
\begin{adjustbox}{max width=\textwidth}
\begin{tcolorbox}[colback=gray!5, colframe=black, title=Prompts, fonttitle=\bfseries]

\begin{tcolorbox}[
  colback=white, 
  colframe=black!30, 
  title=Original prompt, 
  coltitle=black!30!black,
  top=2pt,
  bottom=2pt,
  boxsep=2pt,
  width=\textwidth,
  left=2pt, 
  right=2pt
]
\textbf{USER:} Answer the question using the table below. \\ Table Info: \{table\_info\} \\ Table: \{context\} \\ Question: \{question\} 
\end{tcolorbox}

\begin{tcolorbox}[
  colback=white, 
  colframe=black!30, 
  title=CoT prompting - Zero Shot, 
  coltitle=black!30!black,
  top=2pt,
  bottom=2pt,
  boxsep=2pt,
  width=\textwidth,
  left=2pt, 
  right=2pt
]
\textbf{USER:} Answer the question using the table below. Think step-by-step to explain your answer. \\ Table Info: \{table\_info\} \\ Table: \{context\} \\ Question: \{question\} \\
\textbf{ASSISTANT:} Let's think step-by-step [...]
\end{tcolorbox}

\begin{tcolorbox}[
  colback=white, 
  colframe=black!30, 
  title=CoT prompting - Few-Shot (1), 
  coltitle=black!30!black,
  top=2pt,
  bottom=2pt,
  boxsep=2pt,
  width=\textwidth,
  left=2pt, 
  right=2pt
]
\textbf{USER:} Answer the question using the table below. Think step-by-step to explain your answer. Here is an example that you can follow: \{EXAMPLE\} \\ Now, proceed with this problem - Table Info: \{table\_info\} \\ Table: \{context\} \\ Question: \{question\} \\
\textbf{ASSISTANT:} Let's think step-by-step [...]
\end{tcolorbox}

\end{tcolorbox}
\end{adjustbox}
\caption{Prompts used to infer the different models \cite{kojima2023largelanguagemodelszeroshot}. We selected a correct reasoning trace from the Reasoning version of selected model as a few-shot example for instruct models.}
\label{fig:cot_prompt}
\end{figure}

\noindent \textbf{Efficient Test Time-inference scaling techniques.} Large Reasoning Models tends to generate long reasoning traces. This behavior complicates their  application to real-world agentic environment, where the latency is crucial - specifically in Communication Networks. In the meantime, recent research suggested solutions to mitigate the verbosity of LRMs while retaining their performance \cite{liu2025efficientinferencelargereasoning}. Specifically, inference time-scaling methods enable dynamic inference of the model. To evaluate the potential of inferring reasoning models more efficiently on our Net-Reason dataset, we implemented three existing early-stopping methods:

\begin{enumerate}

    \item \textbf{Prompting:} Recent research showed that LRMs can be stopped using explicit prompting \cite{lee2025llmscompresschainofthoughttoken,han2025tokenbudgetawarellmreasoning}. To replicate this method, we prompted models to explicitly instruct them to not be verbose. We introduce two variants, namely: prompting the model at a system and user prompt level. The system prompt used was \emph{"Respond concisely and confidently. Skip validations and over-verification steps."}, while the user prompt was \emph{"Please do not reason extensively, be succinct."}

    \item \textbf{Black-box method (TRACES):} Alternatively, we implemented TRACES \cite{belkhiter2026tracestaggingreasoningsteps}, an algorithm to early-stop the generation of LRMs dynamically based on the type of reasoning steps. Due to the low token-count relatively to challenging reasoning benchmarks (e.g. less than 2,000 \# toks. vs. $\geq 10,000$ for AIME), we fixed $\delta=0.9$ and a window of $w=1$.

    \item \textbf{Gray-box methods (EAT and DEER):} Further, we evaluated two gray-box methods that uses the model's high-level internals as a signal to early-exit the generation of reasoning models, namely: DEER \cite{yang2025dynamicearlyexitreasoning} and EAT \cite{wang2026entropythinkreasoningmodel}. For the two methods implemented, we selected the hyperparameter recommended by the authors. We set $\lambda=0.95$ for DEER, and ($\alpha=0.2$, $\delta=10^-3$) for EAT.
    
\end{enumerate}

\newpage

\subsection{D. Evaluation pipeline} 

Figure \ref{fig:g_eval_prompt} presents the prompt that we used to evaluate the correctness of models on Net-Reason, using the G-Eval library \cite{liu2023gevalnlgevaluationusing}.  

\begin{figure}[h]
\centering
\footnotesize
\begin{adjustbox}{max width=\textwidth}
\begin{tcolorbox}[colback=gray!5, colframe=black, title=G-Eval prompt, fonttitle=\bfseries]

\texttt{correctness\_metric = GEval( \\
    name="Correctness", \\
    criteria="Determine whether the actual output is correct based on the expected output.", \\
    rubric=[ \\
        Rubric(score\_range=(0.0,0.2), expected\_outcome="Factually incorrect."), \\
        Rubric(score\_range=(0.3,0.6), expected\_outcome="Mostly correct."), \\
        Rubric(score\_range=(0.7,0.9), expected\_outcome="Correct but missing minor details."), \\
        Rubric(score\_range=(1.0,1.0), expected\_outcome="100\% correct."), ]
)}

\end{tcolorbox}
\end{adjustbox}
\caption{Prompt used to evaluate the inference on Net-Reason using \texttt{GPT-4o-mini} \cite{openai2024gpt4ocard} - \url{https://deepeval.com/docs/metrics-llm-evals} (see section \emph{Rubric}).}
\label{fig:g_eval_prompt}
\end{figure}

\subsection{E. Evaluation of Agentic pipelines on Net-Reason}

\begin{figure}[h]
    \centering
    \includegraphics[width=0.8\linewidth]{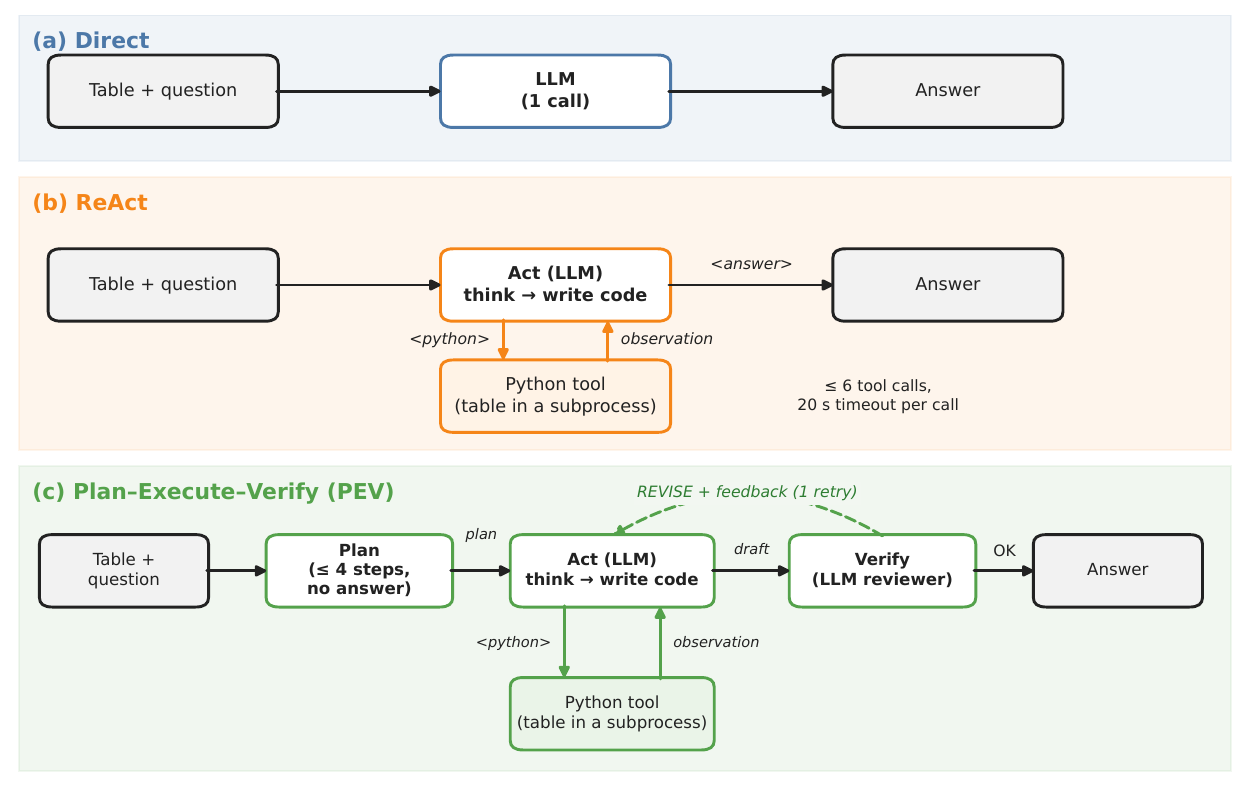}
    \caption{Agentic pipelines evaluated on Net-Reason. \emph{Direct:} single LLM call. \emph{ReAct:} the model alternates between writing Python scripts to analyze the table and reading the output. \emph{PEV:} builds on ReAct, adds a planning step before the tool-call, and a verification step to get feedback.}
    \label{fig:agentic-pipelines}
\end{figure}

\end{document}